\documentclass[10pt,twocolumn,letterpaper]{article}

\usepackage{wacv}              

\usepackage{adjustbox}
\usepackage{algorithm}
\usepackage{algpseudocode}
\usepackage{wacv}
\usepackage{times}
\usepackage{epsfig}
\usepackage[accsupp]{axessibility}
\usepackage{graphicx}
\usepackage{amssymb}
\usepackage{booktabs}
\usepackage{amsmath,multirow,amsfonts,booktabs,xcolor}
\usepackage{array}
\usepackage{textcomp}
\usepackage{stfloats}
\usepackage{url}
\usepackage{verbatim}
\usepackage{enumitem}
\usepackage{colortbl}
\floatstyle{plain}
\restylefloat{figure}

\definecolor{darkgreen}{RGB}{0,200,0}
\usepackage[pagebackref,breaklinks,colorlinks]{hyperref}

\usepackage[capitalize]{cleveref}
\crefname{section}{Sec.}{Secs.}
\Crefname{section}{Section}{Sections}
\Crefname{table}{Table}{Tables}
\crefname{table}{Tab.}{Tabs.}

\def\wacvPaperID{6} 
\def\confName{WACV}
\def\confYear{2024}

\begin{document}

\title{Efficient Domain Adaptation via Generative Prior for 3D Infant Pose Estimation}

\author{
Zhuoran Zhou\textsuperscript{1} \quad
Zhongyu Jiang\textsuperscript{1} \quad
Wenhao Chai\textsuperscript{1} \\
Cheng-Yen Yang\textsuperscript{1} \quad
Lei Li\textsuperscript{2} \quad
Jenq-Neng Hwang\textsuperscript{1} \\
[2mm]
\textsuperscript{1}~University of Washington \quad 
\textsuperscript{2}~University of Copenhagen \\
[2mm]
\tt\small \{zhouz47, zyjiang, wchai, cycyang, hwang\}@uw.edu, lilei@di.ku.dk
}

\maketitle



\begin{abstract}
Although 3D human pose estimation has gained impressive development in recent years, only a few works focus on infants, that have different bone lengths and also have limited data. Directly applying adult pose estimation models typically achieves low performance in the infant domain and suffers from out-of-distribution issues. Moreover, the limitation of infant pose data collection also heavily constrains the efficiency of learning-based models to lift 2D poses to 3D. To deal with the issues of small datasets, domain adaptation and data augmentation are commonly used techniques. Following this paradigm, we take advantage of an optimization-based method that utilizes generative priors to predict 3D infant keypoints from 2D keypoints without the need of large training data. We further apply a guided diffusion model to domain adapt 3D adult pose to infant pose to supplement small datasets.  Besides, we also prove that our method, ZeDO-i, could attain efficient domain adaptation, even if only a small number of data is given. Quantitatively, we claim that our model attains state-of-the-art MPJPE performance of \textbf{43.6} mm on the SyRIP dataset and \textbf{21.2} mm on the MINI-RGBD dataset.
\end{abstract}


\section{Introduction}
\begin{figure}[t]
\centering
  \includegraphics[width=\linewidth]{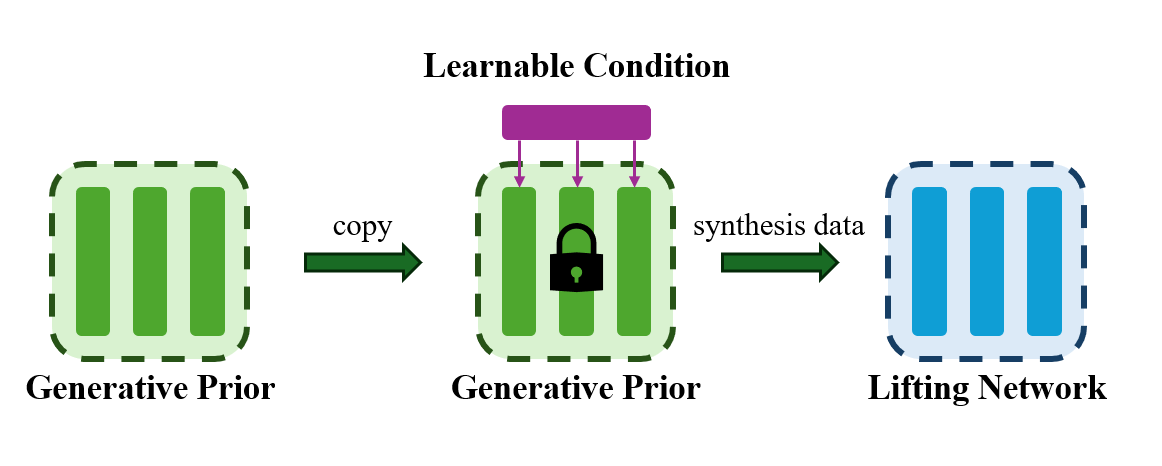}
  \caption{The overall flowchart of our method. Our model aims to adapt a generative prior pre-trained model based on large adult pose data to the infant domain via a controllable branch or fine-tuning. Then, we utilize generative prior in the infant domain to perform optimization work that predicts 3D infant pose from 2D.}
  \label{fig:big_pic}
\end{figure}

3D human pose estimation has been a popular research area these days. Similarly, pose estimation for infants plays an important role in risk assessment and healthcare monitoring\cite{Pierratn741}. However, due to privacy and the difficulty of data collection, public infant pose datasets are rare and limited, and manual labeling is unreliable and expensive. Therefore, it is challenging to train an efficient deep-learning model for infant pose estimation from scratch without sufficient data. To address this limitation, it is natural to think about transferring or tuning an existing adult-based pose estimation model on infant datasets to fully take advantage of similar kinetics of human body pose. Previous work like \cite{syrip} tried to adapt a 2D adult pose detector to the infant domain, but little work has been discussed for 3D infant pose estimation. Therefore, in this paper, we would like to pursue a method that can efficiently predict 3D infant poses even with small infant datasets, by taking advantage of the general kinetic knowledge transferred from an adult adult-based pre-trained model, as the flowchat shown in Figure \ref{fig:big_pic}.

Though learning-based 3D pose estimation models typically learn better features and obtain better performance compared to optimization algorithms, they inevitably require much more data in training for sufficient feature learning. Besides, the majority of human pose estimation learning models suffer from out-of-distribution issues, which make it hard for them to apply in practical scenarios or test data whose characteristics are far from the training data. However, this can easily happen for infant pose as few public datasets are available to support a more general 3D model's training, and cameras in hospitals or healthcare institutions may have different camera settings, leading to an unpredictable domain gap. Fortunately, previous optimization works \cite{bogo2016smplify, thomas2018joint} are proved to be more insusceptible to distribution bias and robust in cross-domain tasks. Moreover, thanks to sophisticated 2D keypoint detectors, two-stage lifting networks are generally of higher accuracy than one-stage networks, which directly predict 3D pose from raw images. In addition, we believe that generation models can be easily trained with few data and be domain-adapted efficiently compared to classic deep-learning models\cite{noguchi2019image}. Therefore, inspired by ZeDO\cite{jiang2023back}, we choose to apply a two-stage optimization-based method, named ZeDO-i, to address the lack of data and out-of-distribution issues under the assistance of generative priors. Given 2D keypoints, our model can iteratively adjust noisy 3D prediction under the constraint of 2D-3D projection and prior distribution learned. As we expected, generative priors learned in the adult domain could be effectively transferred to the infant domain without requiring a lot of data, and the optimization process can cope with challenging test data in reality. Moreover, to simulate the extreme condition of lack of data in small datasets, we also test our model with only 20 and 100 data during adaption and successfully validate our model's ability for efficient domain adaptation. Furthermore, we also introduce a guided diffusion model, which aims to supplement datasets by adapting adult pose to infant pose in order to address data limitation issues and reinforcement diversity. Finally, our method obtains SOTA performance in terms of MPJPE on two infant pose datasets. 
In this paper, we make the following contributions: 

\begin{itemize}
    \item We propose an optimization-based method using generative priors for 3D infant pose estimation. We attain SOTA performance on MINI-RGBD\cite{mini} and SyRIP\cite{syrip}. We also claim that our model can achieve efficient domain adaptation even with a small number of data.
   
    \item We introduce a condition-guided diffusion model which can adapt adult human keypoints to similar infant keypoints for data augmentation purposes and further enhance performance.
\end{itemize}


\section{Related Work}

\subsection{3D Human Pose Estimation}
3D Human Pose Estimation is one of the fundamental tasks in computer vision and is crucial to many downstream tasks, including Human Tracking\cite{andriluka2018posetrack, wang2020combining}, Action Recognition\cite{zhou2021hierarchical, yang2022unsupervised, duan2022revisiting,qi2022weakly}, Motion and Gait Analyses\cite{howe1999bayesian, jiang2022golfpose,zhao2023survey}, and so on. There are three main approaches to realizing the 3D human pose estimation: optimization-based, 2D-3D lifting~\cite{zhang2023mpm,liu2023posynda}, and image-based methods. 

Optimization-based methods are not limited by any training dataset and are good at in-the-wild inference. However, the performance of previous optimization-based methods\cite{bogo2016smplify, tang2018joint, muller2021SMPLify-XMC} is commonly worse than the performance of training-based networks. 2D-3D lifting methods follow a two-stage pipeline requiring a separate 2D human pose estimation model and  a lifting network to map 2D human poses to 3D human poses in single frames or short sequences. Pavllo \etal~\cite{pavllo20193d} apply dilated temporal convolution to enhance 3D pose estimation for unlabeled videos in a semi-supervised method. Zhao \etal~\cite{zhao2019semantic} design a novel graph convolution and take advantage of a graph convolution network (GCN) to learn inter-joint features and local and global relationships in a structured graph. On the other hand, image-based methods focus on directly regressing 3D human poses from RGB images. Kolotouros \etal~\cite{kolotouros2019learning} introduce SPIN (SMPL oPtimization IN the loop) by using a CNN to extract features from a cropped-out human image and regress the SMPL~\cite{SMPL} parameters with the help of an optimization-based pose estimation pipeline to conduct semi-supervised learning. However, all the learning-based methods suffer from the use of small datasets in Infant Pose Estimation tasks. In this paper, we focus on how to conduct 3D infant pose estimation with limited data.

\subsection{Infant Pose Estimation}
Infant pose estimation, which aims at predicting 2D and 3D keypoints of infants in image and world coordinates, can lead to useful downstream tasks such as infant action recognition \cite{yang2022unsupervised, huang2023posture} or motion analysis \cite{chambers2020computer, huang2023sym}. Hesse \etal\cite{mini} are the first to present the MINI-RGBD dataset, which enables the experiment on 2D infant pose estimation. Subsequently, Huang \etal\cite{syrip} propose a hybrid synthetic and real infant pose (SyRIP) dataset based on SMIL \cite{ma2021smil} with annotated 2D keypoints. Following the 2D infant pose estimation, the mainstream of 3D infant pose estimation works on RGB-D data. Wu \etal\cite{wu93} measure infant movements by combining 2D keypoints and matching depth images collected by Kinect. Li~\cite{li2020three} continues using the same pipeline but correcting depth information for a better matching between image and depth. However, Kinect may cause depth ambiguity if joints are occluded, and depth images are not always available in the infant monitor system. In ~\cite{ellershaw3d}, the author uses a 2D pose estimation model and a 3D lifting network pretrained on the adult dataset and fine-tuned on the infant dataset. Though this model achieves rather good performance on the MINI-RGBD dataset, it is basically learning-based and hard to adapt to more realistic data due to the domain gap. From our experience in human pose estimation, predicting 3D keypoints from 2D keypoint detection is easier than the one-stage method predicting 3D joints directly from raw images.


\section{Methods}
Our model primarily consists of a diffusion model to learn the prior, and an optimization algorithm to iteratively adjust 3D pose prediction. Additionally, we apply a condition-guided diffusion model for pose data augmentation.  
We demonstrate the method as followings: background of diffusion model in section~\ref{sec:method_diffusion}, generative prior model in section~\ref{sec:prior model}, optimization algorithm in~\ref{sec:opt}, its controllable adaptation variant in~\ref{sec:ctrl} and condition-guided data augmentation in~\ref{sec:aug}.   

\begin{figure*}[t]
\centering
  \includegraphics[width=\textwidth]{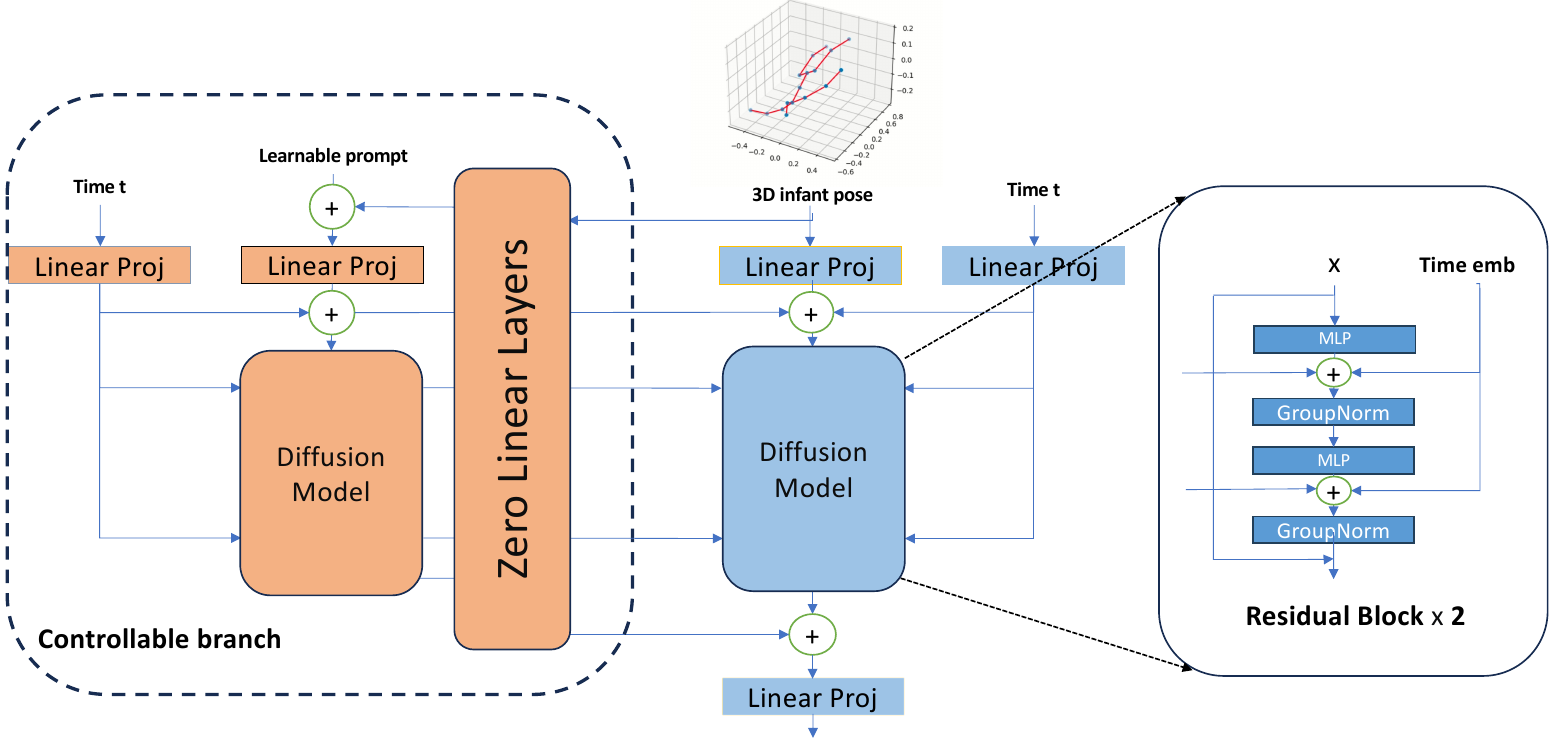}
  \caption{\textbf{Model architecture}. The modules excluded by the dotted box comprise our proposed score-matching-based prior learning model, and the modules inside the dotted box are the controllable branch used in one of our adaptation strategies. The prior learning model consists of MLP layers and two residual blocks displayed on the right side. The controllable branch takes a learning prompt as inputs, copies the weights from the prior model and only updates the copied weights during adaptation, while the original prior learning model is kept frozen. For convenience, we paint all frozen layers in controllable adaptation as blue and all updated layers as brown.}
  \label{fig:arch}
\end{figure*}

\subsection{Preliminaries of Diffusion Model}
\label{sec:method_diffusion}
Before introducing our diffusion model as the generative prior learning model, we briefly discuss its background for clarity. Diffusion models~\cite{ho2020denoising,song2020score} are popular generation models used in tasks like image generation~\cite{ramesh2021zero}, image inpainting~\cite{lugmayr2022repaint}, editing~\cite{cao2023difffashion} and so on. During training, the diffusion model iteratively adds Gaussian noise to an image relative to a timestamp $t$ to the inputs and tries to recover inputs from a noise image in the reverse process. In this paper, we use Score-Matching-Network(SMN)~\cite{song2020score} as our prior learner. SMN aims to train a score network $s_{\theta}(x)$ to approximate gradients of log probability of a score function $p_{\theta}(x)$, expressed as $s_{\theta}(x) \approx \triangledown_{x} \log{p_{\theta}(x)}$, so the loss is generally represented as 

\begin{equation}
    E_{p(x)} \| \triangledown_{x} \log{p_{\theta}(x)} -  s_{\theta}(x)\|^{2}_2.
\end{equation}

\subsection{Infant Pose Prior Model}
\label{sec:prior model}
For pose estimation tasks on small datasets, learning-based deep-learning models suffer from out-of-distribution issues and insufficient resources to extract reliable features. Built upon the work of ZeDO~\cite{jiang2023back}, we also propose to use an optimization-based method to predict 3D keypoints from 2D keypoints along with a score-matching network diffusion model(SMN)~\cite{song2020score} as our prior learner. Our final architecture is illustrated in \ref{fig:arch}. The modules excluded by the dotted bounding box comprise our proposed prior learning model, which takes root-relative infant keypoints, sized $B \times J \times 3$, and randomly sampled noise timestamp $t$ as inputs, where $J$ is the number of joints. The embedding layers are simple linear projection layers, which lift input dimension to $B \times 1024$, and then sum them up. Further, the embedding goes through the Score-Matching-Network diffusion model consisting of two residual blocks as backbones. Each of the residual blocks contains two residual-connected MLPs. The last output projection layer projects the feature back to pose joints. With the generative priors, our method can denoise a noisy 3D pose in the optimization stage if it violates the kinematic rules of infant poses.

\subsection{Optimization Algorithm}
\label{sec:opt}
Given a 2D infant pose and the intrinsic parameter, ZeDO-i first tries to compute the ray vectors emitted by the camera and initializes the predicted 3D keypoints on the rays to minimize 3D-2D projection errors. Further it activates the generation model to adjust the noisy 3D pose prediction based on its prior knowledge. After each adjustment, 3D keypoints may be off the rays, and the model again moves them onto the rays in the shortest distance. Our method runs this iteration 1000 times to iteratively achieve a reasonable 3D pose under a 3D-2D projection constraint. 
In experiments, we find that a pseudo intrinsic parameter which has a focal length of 2000 and a camera center equal to the image center also functions so one could apply it in practical cases.

In details, we first define ray vectors $\hat{V_{ray}}$ emitted from the camera using 2D keypoints $X_{2D}$ and real or pseudo intrinsic parameter $K$,  in which focal length is always 2000 and the principal point is the image center point.
Then we randomly choose a training 3D pose $X_{3D_init}$ and use an Adam optimizer~\cite{kingma2014adam}, which helps us find an appropriate rotation $R_{o}$ and translation $T_{0}$ such that $ \|K(R_{0}X_{3D_init} + T_{0}) - X_{2d}\|_2$ is minimized. With $T_{0}$ known, we set all initial 3D keypoints on the rays with depth equal to $T_{0}$, and supposedly this 3D pose has zero projection error with the 2D ground truth. Next, we start $T=1000$ times of optimization steps in which we first move 3D kyepoints to corresponding rays in the shortest distance if they are off the rays and then the prior model is used to adjust the noisy pose based on the prior distribution it learns. 

In evaluation, we find that a noise level $t=(0,0.1]$ works the best, and we also observe that performance is heavily dependent on the initial depth distance assigned. As the training data are all root-relative, the prior model may cause depth ambiguity if we don't limit depth in the first few steps. In practice, we get the lowest error when forcing the depth $T$ unchanged in the first 950 iterations and opening the constraint in the remaining $50$.

\subsection{Controlling Branch for Domain Adaptation}
\label{sec:ctrl}
As the kinematics of infants and adults are similar, transferring a pre-trained adult pose model to the infant domain would intuitively boost the performance. Considering that directly fine-tuning a model trained on a huge amount of data to a small dataset may lead to overfitting, we propose a method inspired by Control-Net \cite{zhang2023adding} to manipulate the adaption process of the generative priors. As shown in the left-side dotted box in Figure \ref{fig:arch},  we duplicate the weights of the prior model to the controllable branch. Like how Control-Net sets the condition, We set a learnable prompt with the same size as 3D pose as the controlling inputs and connect the prior model and controllable branch with a few zero linear layers, which are fully initialized as zero weights. Then internal embeddings are added back to the prior model before and after every residual block. During adaptation, all layers of the prior model have to be kept frozen, and only the controllable branch is open to weight update. In the experiment section, we will compare its performance with two other adaption strategies: fine-tuning a pre-trained adult prior model and training a prior model in the infant domain only. 

\subsection{Condition Guided Diffusion Model}
\label{sec:aug}
\begin{figure}[t]
\centering
  \includegraphics[width=0.5\textwidth,scale=0.5]{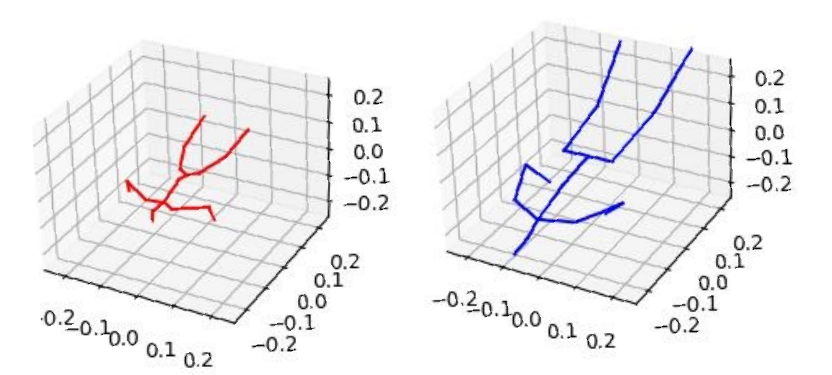}
  \caption{Left: augmented infant pose. Right: h36m adult pose. Our augmentation model converts the adult pose to a similar infant pose by adjusting scales and kinematic features like bone length without altering actions much. }
  \label{fig:augvis}
\end{figure}

If the only available data are too few to be used, one could convert the resourceful adult poses to an infant-like pose as data augmentation. To fit the augmented data to the kinematics of infants, we trained a score matching network diffusion model which takes in both adult and infant 3D poses along with two $1\times 1000$ sized learnable condition tokens to represent whether the pose belongs to adults or infants. The architecture is similar to the prior model we used. We hope that the diffusion model would implicitly learn features like bone lengths and bone angles for two different domains and know how to distinguish their distribution. During inference, we ask the diffusion model to generate corresponding infant poses based on the given input adult poses, so the model adjusts the scale and angle according to the implicit knowledge of the pose prior, yet still keeps the pose semantic meanings, such as actions, as shown in Figure \ref{fig:augvis}. We prove that adding these challenging poses enhances diversity in the ablation study.

\begin{figure*}[t]
\centering
  \includegraphics[width=\linewidth]{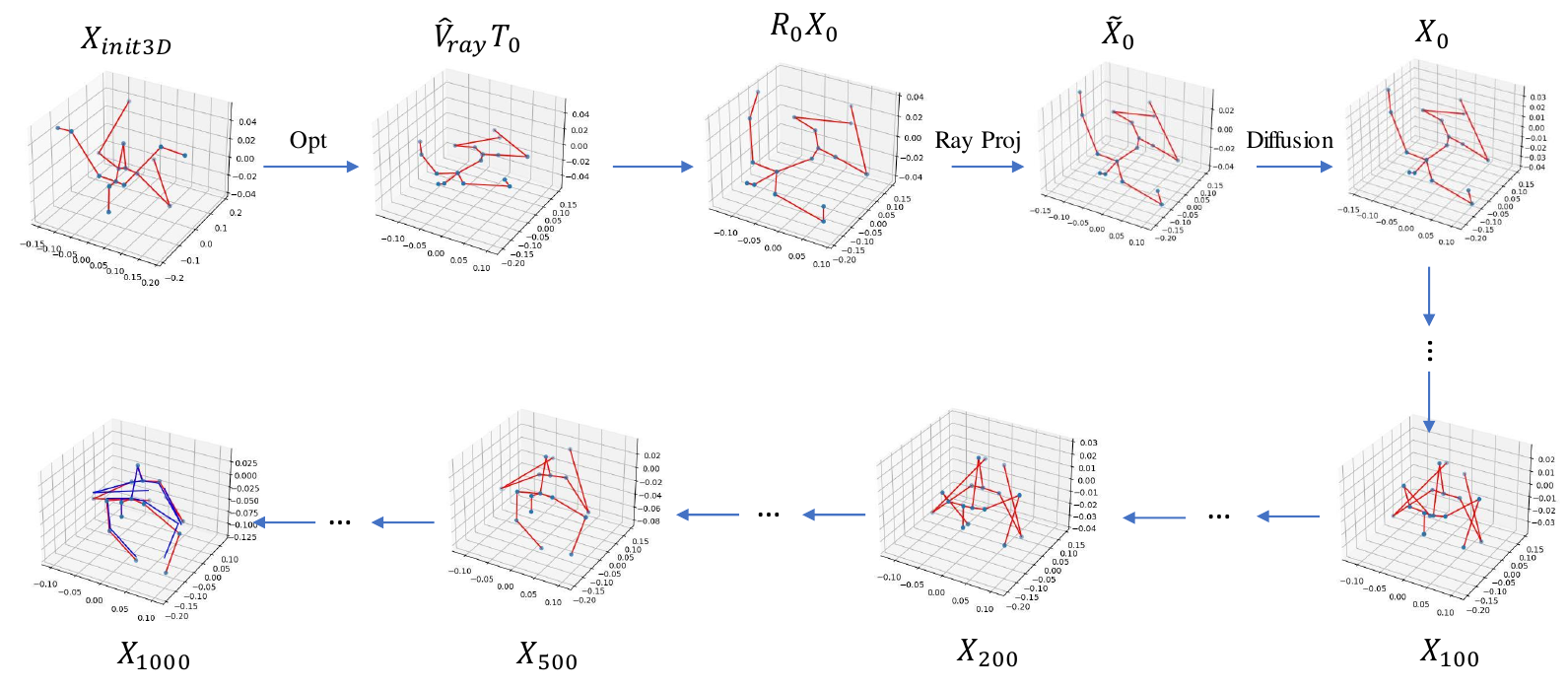}
  \caption{Visualization of the optimization algorithm. $\hat{V_{ray}}$ ray vectors are first calculated.  $T_{0}$ and $R_{0}$ are then found via an optimizer, so the initial pose is $T_{0}$ in depth on the rays. We then run the optimization algorithm 1000 times. In each step, keypoints are moved toward the rays, represented by $\tilde{X_{0}}$ in the figure, and are also sent to the diffusion model to adjust the pose to get $X_{0}$. In the last step $X_{1000}$ the ground truth in blue is quite close to our prediction in red.}
  \label{fig:diff}
\end{figure*}

\section{Experiments}

\subsection{Datasets}
We conduct our experiments on MINI-RGBD\cite{mini} and SyRIP\cite{syrip}, two public infant datasets with 2D-3D pose pairs. For pre-trained adult prior model and condition-guided diffusion model, we take advantage of Human3.6m\cite{H36M}.

\noindent\textbf{MINI-RGBD} includes 12 sequences of data, in total 12000 synthetic infant images, and also provides their 25 joint 2D and 3D keypoint pairs. We train on the first 9 sequences and test on the rest 3,  following the 17-keypoint definition of Human3.6m.

\noindent\textbf{SyRIP} includes a diverse set of 700 real and 1000 synthetic infant images, generated by fitting SMIL~\cite{ma2021smil} models to real images, supplemented with additional variants to the SMIL shape and pose parameters. Later, in total of 700 weak ground truth of 3D keypoints in ~\cite{huang2023sym} were manually corrected and made available. In this paper, we also train on their 600 weak ground truth 3D labels and test on the 100 real images.

\noindent\textbf{Human3.6m (H36m)}\cite{H36M} is a single-frame 3D human pose benchmark, containing about 3.6 million 2D-3D human pose pairs. This dataset was collected in an indoor setting, consisting of 17 various actions. As this adult 3D pose dataset includes more actions and diversity than infant 3D poses, we intend to transfer its 3D poses to the infant domain for data augmentation.

\subsection{Implementation Details}
We pre-train our adult prior model on Human3.6m\cite{H36M}, which includes millions of adult 3D pose data, and further train all three adaptation strategies for $5000$ epochs with a learning rate of $2\times10^{-4}$. During training, we set the total diffusion step as $1000$ with a uniform noise level of $\left[0, 1.0\right]$. We use the Adam Optimizer with a batch size of $5000$. 

In inference, we choose a noise level $t$ in $\left(0,0.1\right]$ and run the optimization for 1000 iterations. We keep the depth distance unchanged in the first $950$ iterations.

The guided diffusion data augmentation diffusion model shares the same training configuration as the prior model. We choose a noise level in the range $\left(0,1.0\right]$ and only run $100$ iterations for the diversity of augmented data. In experiments, we add 600 augmented data to SyRIP and 4000 to MINI-RGBD.

\subsection{Experiment Results}
In this section, we first compare our method's results to the previous SOTA in terms of MPJPE. In addition, we also test if our method can be efficiently adapted to the infant domain with 20 and 100 data only in order to simulate extreme situations.  Further, we evaluate all domain adaptation strategies of our model including the controllable adaptation method~(CA), fine-tuning from the adult-based diffusion model~(FT), and training from scratch on infant data to seek the best adaptation approach.

\begin{table}[t]
\centering
\resizebox{0.6\linewidth}{!}{%
\begin{tabular}{l | c }
\toprule
 Methods  & MPJPE~($\downarrow$)\\
\midrule
  Kolotouros~\etal~\cite{kolotouros2019learning} & 105.8 \\
  Liu~\etal~\cite{liu2021heuristic} &  97.2\\
  Liu~\etal(Finetuned)  & 78.3  \\
\midrule
  ZeDO-i (GT)  & \textbf{43.6} \\
  ZeDO-i (DT)   & 47.7 \\
 

\bottomrule
\end{tabular}
}
\caption{3D infant pose estimation results on SyRIP dataset under 12 joints setting. For a fair comparison, we list the performance of ZeDO-i with both estimated 2D keypoints and ground truth 2D keypoints.Estimated 2D keypoints and other method performance are provided in \cite{huang2023sym}.}
\label{tab:syrip}
\end{table}

\subsection{Results on SyRIP}
Similar to previous works, we have the same training and testing sets as~\cite{huang2023sym} with only 12 keypoints of limbs for fair comparison. Observed from Table~\ref{tab:syrip}, our method clearly achieves the SOTA performance 
even with only 20 training data. Moreover, as shown in Table \ref{tab:few_data_syrip}, we observe that the controllable adaptation approach achieves better results than fine-tuning when the data number is small, therefore controllable adaptation is more suitable for limited data in such more practical and diverse scenarios. Both adaptation and fine-tuning from the adult domain are better than training from scratch, indicating that knowledge from the adult domain is necessary.

\begin{table}[t]
\centering
\resizebox{0.6\linewidth}{!}{%
\begin{tabular}{l | c }
\toprule
 Methods  & MPJPE~($\downarrow$)\\
\midrule
  Hesse \etal\cite{hesse} & 44.9 \\
  Ellershaw$^*$~\etal\cite{ellershaw3d} &  34.2\\
  Ellershaw~\etal  &  28.5\\
  \midrule
  ZeDO-i &  \textbf{21.2}\\
 

\bottomrule
\end{tabular}
}
\caption{3D infant pose estimation results on MINI-RGBD under 16 joints setting. We list the best performance among the three strategies. $^*$ denotes w/o adult pre-training. We evaluate 16 keypoints to keep aligned with the setting of the previous SOTA.}
\label{tab:mini}
\end{table}

\subsection{Results on MINI-RGBD}
For a fair comparison with previous works\cite{ellershaw3d}, we follow their keypoint definitions and show the results in Table \ref{tab:mini}. Our method beats all previous SOTA to a great extent.
\\
Besides, as shown in Table \ref{tab:few_data_mini}, direct fine-tuning adult pre-trained model on MINI-RGBD attains lower error, which is different from SyRIP. We suspect that the discrepancy between training and testing sets leads to this observation, as MINI-RGBD is full of synthetic images with rather less discrepancy compared to the SyRIP dataset. Moreover, we also include the results of 16 keypoints like the previous SOTA, showing that our model is already comparable to the previous SOTA with only 100 training data.

\section{Ablation Studies}

\begin{table}[t]
\centering
\resizebox{0.8\linewidth}{!}{%
\begin{tabular}{l | c | c | c}
\toprule
 Datasets   & CA & FT & From Scratch\\
\midrule
  SyRIP~($S$=20) & \textbf{67.8} &69.4 & 72.3\\
  SyRIP~($S$=100) & \textbf{56.4} & 60.8 & 60.6 \\
  SyRIP(GT)   &49.4 & \textbf{47.7} & 54.0\\
  SyRIP~(augmented)   & 45.5 & \textbf{43.6} &48.9\\
  

\bottomrule
\end{tabular}
}
\caption{MPJPE performance of different strategies on SyRIP. The controllable adaptation approach achieves better performance than the other two approaches when the data number is small.}
\label{tab:few_data_syrip}
\end{table}


\begin{table}[t]
\centering
\resizebox{0.85\linewidth}{!}{%
\begin{tabular}{l | c | c | c| c}
\toprule
 Datasets   & CA & FT & From Scratch & Best($J$ = 16)\\
\midrule
  MINI-RGBD~($S$=20) & 38.7 & 36.8 & \textbf{36.4} & 34.6\\
  MINI-RGBD~($S$=100) & 34.8 & \textbf{31.9} & 33.7 & 29.4\\
  MINI-RGBD Full & 25.5 & \textbf{24.1} &27.4 & 22.8\\
  MINI-RGBD~(Augmented)  &20.7 &\textbf{19.9} &21.3 & 21.2\\


\bottomrule
\end{tabular}
}
\caption{MPJPE performance of different strategies on MINI-RGBD. Here, we not only evaluate all 17 keypoints according to H36M's keypoint definition but also list their performance in 16 keypoints for the convenience of fair comparison with previous works.}
\label{tab:few_data_mini}
\end{table}

\begin{figure}[t]
\centering
  \includegraphics[width=0.85\linewidth]{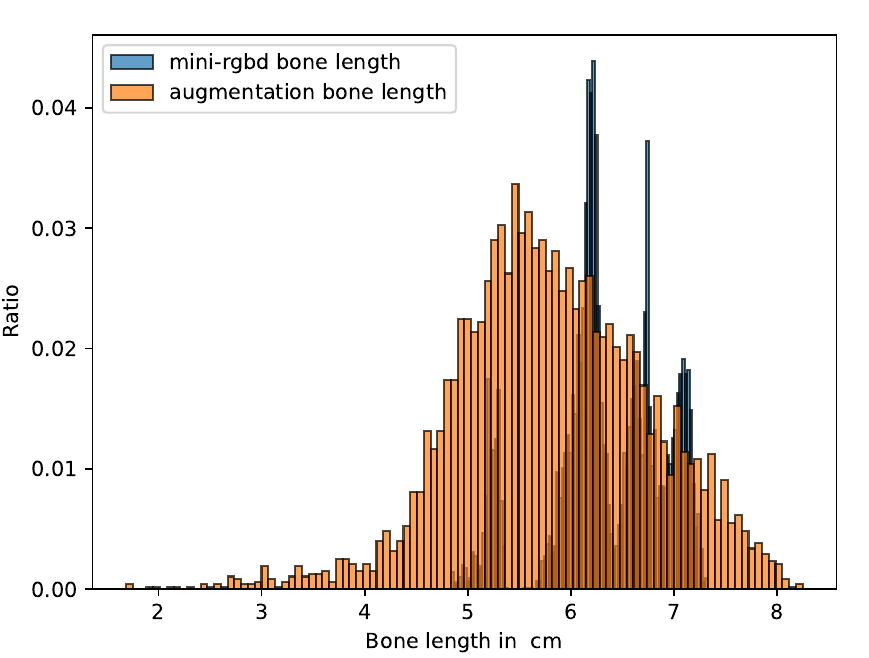}
  \caption{The data distribution of bone length in the augmented dataset is better than the original MINI-RGBD. Our augmented dataset spans over a wider range of bone lengths.}
  \label{fig:bl1}
\end{figure}

\begin{figure}[t]
\centering
  \includegraphics[width=0.85\linewidth]{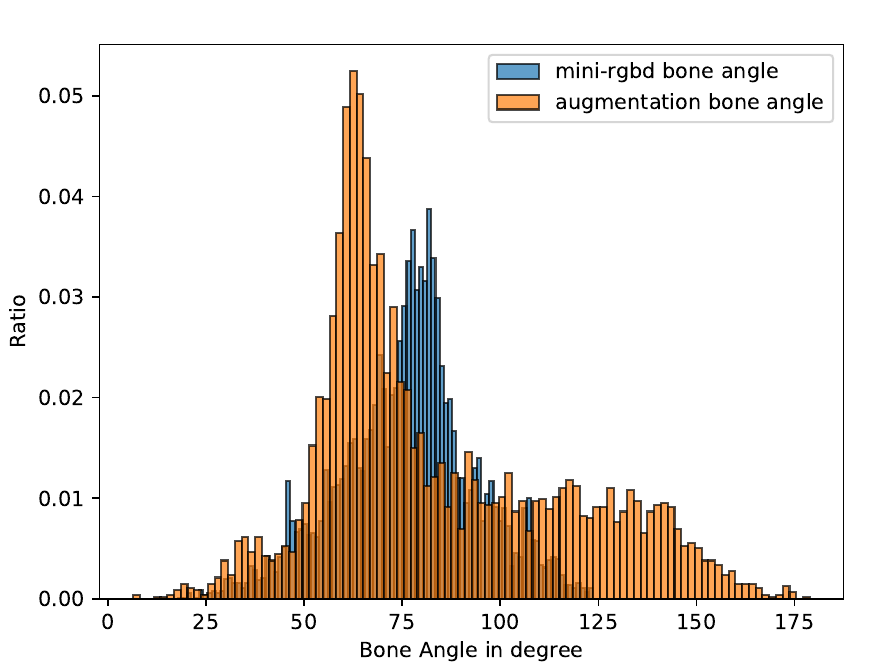}
  \caption{The data distribution of bone angle in the augmented dataset is better than the original MINI-RGBD. Our augmented data have a wider range of bone angles.}
  \label{fig:angle1}
\end{figure}

\begin{figure*}[t]
\centering
  \includegraphics[width=\linewidth]{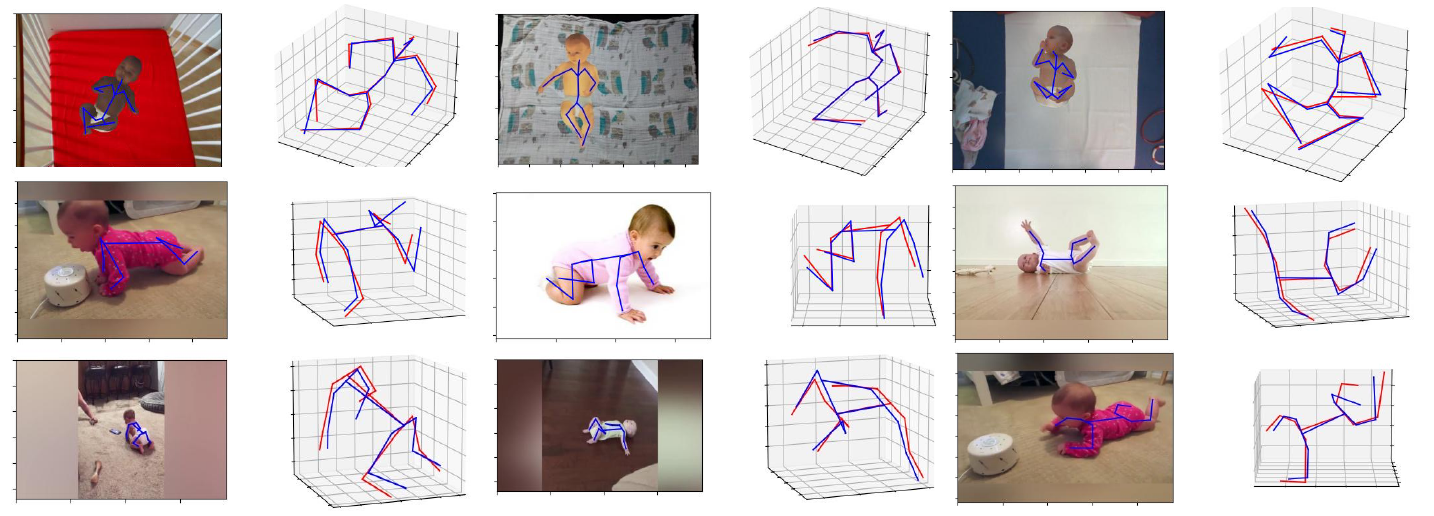}
  \caption{Visualizations of 2D ground truths and our 3D predictions on MINI-RGBD(Top Line) and SyRIP(Middle and Bottom Line). Our 3D predictions are colored in red, and the ground truth are in blue. The shapes and poses in general are well aligned.}
  \label{fig:syrip_vis}
\end{figure*}

\begin{figure}[t]
\centering
  \includegraphics[width=\linewidth]{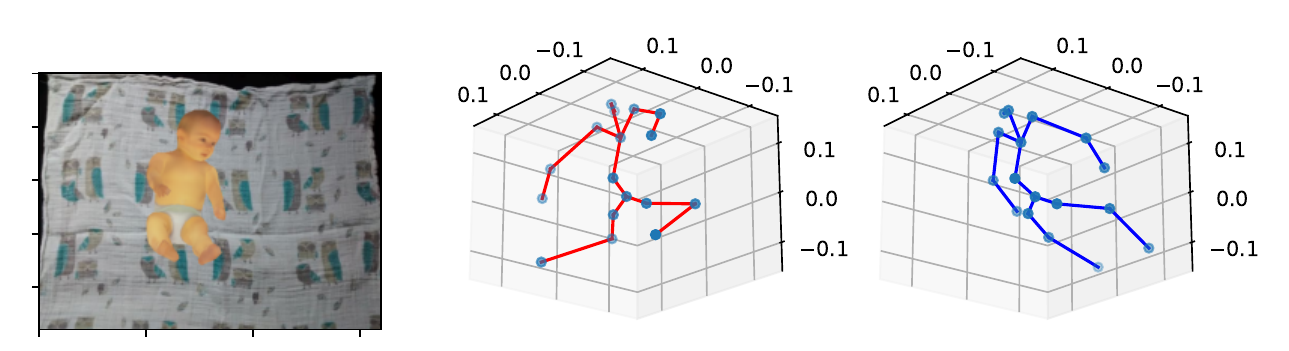}
  \caption{One of the failure examples achieves the highest MPJPE of 160mm in MINI-RGBD. Left side: our prediction. Right side: ground truth.}
  \label{fig:failure}
\end{figure}

\subsection{Data Augmentation Diversity}
\label{sec:ablation_aug}
In this section, we demonstrate how our condition-guided data augmentation method enhances the diversity in the MINI-RGBD dataset as its data are all synthesized and narrow-distributed. We analyze bone lengths and bone angles of the original dataset and our augmented data. As shown in Figure \ref{fig:bl1}, we randomly choose one bone and compare their lengths. Our augmented bone length spans over a wider range of scales. Similarly, we show comparisons of bone angles in Figure \ref{fig:angle1} and get the same conclusion. 
Tables \ref{tab:few_data_mini} and \ref{tab:few_data_syrip} also justify this conclusion quantitatively. 

\subsection{Comparison with 3D pose estimation model}
To prove the efficiency of our two-stage optimization method, we compare its performance with other classic learning-based 3D pose estimation models widely applied in human pose estimation tasks. We train all the models from scratch on infant datasets without data augmentation.

As shown in Table \ref{tab:compare}, our method outperforms the classic 2D-3D lifting human pose estimation model, VideoPose3D\cite{pavllo20193d}, which further proves our claim that the proposed optimization method can better fit the task of small-dataset domain adaption in 3D pose estimation than other learning-based models.

\begin{table}[t]
\centering
\resizebox{0.6\linewidth}{!}{%
\setlength{\tabcolsep}{0.1cm}
\begin{tabular}{l | c | c }
\toprule
 Datasets   & ZeDO-i  & VideoPose3D\cite{pavllo20193d}\\
\midrule
  SyRIP   &\textbf{49.6} & 145.7\\
  MINI-RGBD   &\textbf{27.4} & 106.7\\


\bottomrule
\end{tabular}
}
\caption{MPJPE performance of different pose estimation models trained from scratch without augmentation. Clearly, ZeDO-i is more suitable for infant small dataset than learning-based models.}
\label{tab:compare}
\end{table}


\section{Limitation}
Though our method achieves impressive performance in small datasets like infant 3D pose, it still needs accurate 2D keypoints. Additionally, the prediction results of our method also depend on the depth distance $T_{0}$ defined in initialization since we find that the generation model only learns root-relative priors with little knowledge of spatial depth. 
Besides, like all optimization works aiming to minimize 2D-3D projection error, our method may also suffer from one-to-many mappings. For example, we show one failure example in MINI-RGBD in Figure \ref{fig:failure}. Here our model fails to predict the correct $T$ of 3D keypoints in spite of the matched 2D projections. We calculated the median MPJPE error which is 4mm lower than the mean, which implies that these extreme outlines are very rare.


\section{Conclusion}
 We propose an optimization method which applies generative priors of the infant pose to predict 3D infant keypoints. We show that our method achieves SOTA on MINI-RGBD and SyRIP and attains efficient domain adaptation using a small amount of data. Besides, we compare three training strategies for our model, in which fine-tuning an adult pre-trained generative model seems more efficient for MINI-RGBD and the whole SyRIP dataset, but the controllable adaptation version performs better in SyRIP when only 20 and 100 data are available. We also introduce a condition-guided diffusion model which enhances the kinematic diversity and boosts overall results. In general, we state that our method fits the small-dataset 3D infant pose estimation very well and attains outstanding performance.

\newpage
{\small
\bibliographystyle{ieee_fullname}
\bibliography{main}
}

\end{document}